\documentclass[conference]{IEEEtran}
\IEEEoverridecommandlockouts
\usepackage{cite}
\usepackage{amsmath,amssymb,amsfonts}
\usepackage{algorithmic}
\usepackage{graphicx}
\usepackage{textcomp}
\usepackage{xcolor}
\usepackage{algorithm}
\usepackage{placeins}
\usepackage{verbatim}

\def\BibTeX{{\rm B\kern-.05em{\sc i\kern-.025em b}\kern-.08em
    T\kern-.1667em\lower.7ex\hbox{E}\kern-.125emX}}
\begin{document}

\title{D-SPEAR: Dual-Stream Prioritized Experience Adaptive Replay for Stable Reinforcement Learning in Robotic Manipulation\\

}

\author{
\IEEEauthorblockN{Yu Zhang}
\IEEEauthorblockA{\textit{School of Computer Science} \\
\textit{University of Galway}\\
Galway, Ireland \\
Y.Zhang34@universityofgalway.ie}
~\\
\and
\IEEEauthorblockN{Karl Mason\textsuperscript{*}}
\IEEEauthorblockA{
\textit{School of Computer Science} \\
\textit{University of Galway} \\
Galway, Ireland \\
karl.mason@universityofgalway.ie
}
\IEEEauthorblockA{*Corresponding author}
}

\maketitle

\begin{abstract}
Robotic manipulation remains challenging for reinforcement learning due to contact-rich dynamics, long horizons, and training instability. 
Although off-policy actor--critic algorithms such as SAC and TD3 perform well in simulation, they often suffer from policy oscillations and performance collapse in realistic settings, partly due to experience replay strategies that ignore the differing data requirements of the actor and the critic.

We propose D-SPEAR: Dual-Stream Prioritized Experience Adaptive Replay, a replay framework that decouples actor and critic sampling while maintaining a shared replay buffer. 
The critic leverages prioritized replay for efficient value learning, whereas the actor is updated using low-error transitions to stabilize policy optimization. 
An adaptive anchor mechanism balances uniform and prioritized sampling based on the coefficient of variation of TD errors, and a Huber-based critic objective further improves robustness under heterogeneous reward scales.

We evaluate D-SPEAR on challenging robotic manipulation tasks from the robosuite benchmark, including Block-Lifting and Door-Opening. 
Results demonstrate that D-SPEAR consistently outperforms strong off-policy baselines, including SAC, TD3, and DDPG, in both final performance and training stability, with ablation studies confirming the complementary roles of the actor-side and critic-side replay streams.
\end{abstract}


\section{Introduction}

Robotic manipulation remains difficult for reinforcement learning because contact-rich interactions and long horizons often produce unstable training dynamics~\cite{tang2025deep}. Although off-policy actor--critic methods such as SAC and TD3~\cite{SAC1,TD3} perform well in simulation, they can still exhibit high variance, policy oscillations, and late-stage collapse in realistic manipulation settings~\cite{christmann2024benchmarking}.

Experience replay improves off-policy learning through data reuse~\cite{ER}, but uniform replay treats all transitions as equally useful. In robotic manipulation, replay buffers are often dominated by low-information samples, while rare but informative transitions are under-utilized. Prioritized Experience Replay (PER) addresses this with TD-error-based sampling~\cite{PER}, yet high-TD samples can destabilize continuous actor--critic learning by injecting unreliable value gradients into policy updates~\cite{APER2023actor}. This reveals a mismatch: the critic benefits from high-error samples, whereas the actor benefits from stable, low-error ones~\cite{ACarchi}.

To address this mismatch, we propose \textbf{D-SPEAR}, which decouples actor and critic sampling while keeping a shared replay buffer. D-SPEAR combines critic-side prioritization, actor-side low-error replay, an adaptive anchor mechanism based on TD-error variation, and a Huber-based critic objective~\cite{Huber2020equivalence}. This design provides a simple yet effective mechanism for balancing value learning and policy stability, particularly in robotic manipulation tasks where contact events and sparse rewards often produce highly uneven TD-error distributions.

We evaluate D-SPEAR on the \textit{robosuite} Lift and Door tasks. It consistently outperforms SAC, TD3, and DDPG~\cite{DDPG} in final return and training stability, while ablations confirm the complementary roles of the two replay streams~\cite{robosuite2020}. These results indicate that separating replay strategies for actor and critic updates can improve learning stability in contact-rich manipulation environments. More broadly, the proposed framework illustrates how replay mechanisms can be adapted to actor–critic optimization roles.

\subsection{Reinforcement Learning for Robotic Manipulation}

Reinforcement learning has been widely applied to robotic manipulation, including grasping, lifting, and articulated object interaction~\cite{levine2016end,kalashnikov2018scalable,tang2025deep}. These tasks are typically formulated as Markov Decision Processes (MDPs), which yield transition tuples $(s_t,a_t,r_t,s_{t+1})$ stored in replay buffers for off-policy learning. Because manipulation involves contact-rich dynamics, delayed rewards, and long horizons, model-free methods remain attractive but often suffer from unstable training.

\subsection{Soft Actor--Critic for Continuous Control}

Among off-policy actor--critic methods, Soft Actor-Critic (SAC) has emerged as a strong baseline for complex continuous control tasks~\cite{SAC1}. In addition, Twin Delayed Deep Deterministic Policy Gradient (TD3) is another widely used off-policy actor--critic algorithm and serves as an important baseline in our experiments. SAC is derived under the maximum entropy reinforcement learning framework, which augments the standard expected return objective with an entropy regularization term. This formulation encourages policies that maximize both task performance and stochasticity, leading to improved exploration and robustness:
\begin{equation}
J(\pi) = \mathbb{E}_{\tau \sim \pi} \left[ \sum_{t=0}^{\infty} \gamma^t \left( r(s_t, a_t) + \alpha \mathcal{H}(\pi(\cdot \mid s_t)) \right) \right],
\end{equation}
where $J(\pi)$ denotes the maximum entropy objective for policy $\pi$, $\tau = (s_0, a_0, s_1, a_1, \ldots)$ represents a trajectory sampled under policy $\pi$, and $\gamma \in (0,1)$ is the discount factor. The term $r(s_t,a_t)$ denotes the reward received at time step $t$, $\mathcal{H}(\pi(\cdot \mid s_t)) = -\mathbb{E}_{a \sim \pi}[\log \pi(a \mid s_t)]$ is the entropy of the policy at state $s_t$, and $\alpha > 0$ is the temperature parameter that controls the trade-off between reward maximization and entropy. In practice, $\alpha$ is automatically tuned following the standard SAC implementation, and the same setting is adopted in D-SPEAR.

SAC employs an off-policy actor--critic architecture with clipped double Q-learning to mitigate overestimation bias. The critic is trained to minimize a soft Bellman residual, while the actor is updated by minimizing the Kullback--Leibler divergence between the policy and an energy-based distribution induced by the Q-function~\cite{SAC2}. This combination enables SAC to achieve high sample efficiency and stable performance in high-dimensional, nonlinear control problems. As a result, SAC has been widely adopted in robotic manipulation tasks, where efficient exploration and robustness to modeling errors are critical~\cite{10474399}.

\subsection{Experience Replay, Prioritization, and Robust Objectives}

Experience replay (ER) is a fundamental mechanism that enables sample-efficient off-policy learning~\cite{ER}. By storing transitions collected through interaction with the environment and reusing them for multiple updates, ER breaks temporal correlations in training data and improves data efficiency. Standard ER relies on uniform sampling, implicitly assuming that all stored transitions contribute equally to learning. In robotic manipulation tasks, however, replay buffers are often dominated by low-information transitions, particularly in settings with sparse rewards or delayed task completion signals.

Prioritized Experience Replay (PER) was introduced to address this limitation by sampling transitions in proportion to their temporal-difference (TD) errors, under the assumption that larger errors indicate higher learning potential~\cite{PER}. While PER has demonstrated clear benefits in value-based methods, its application to continuous actor--critic algorithms can introduce instability. High TD-error transitions often correspond to regions where the critic’s value estimates are inaccurate. In actor--critic methods, policy updates depend on value gradients $\nabla_a Q(s,a)$, typically expressed as
\begin{equation}
\nabla_{\theta} J(\theta) = \mathbb{E}_{s \sim D}\left[\nabla_a Q(s,a)\nabla_{\theta}\pi_{\theta}(s)\right].
\end{equation}
Oversampling transitions with large TD errors can therefore amplify gradient variance during policy updates, which may destabilize the actor.

Beyond sampling strategies, the robustness of actor--critic learning is also influenced by the choice of loss function. Mean squared error objectives are sensitive to outliers and unnormalized rewards, which frequently arise in robotic manipulation environments~\cite{APER2023actor}. Robust alternatives, such as the Huber loss~\cite{huber1992robust}, mitigate this issue by bounding gradient magnitudes under large errors. The Huber loss is defined as
\begin{equation}
\mathcal{L}_{\delta}(x) =
\begin{cases}
\frac{1}{2}x^2, & \text{if } |x| \leq \delta, \\
\delta \left( |x| - \frac{1}{2}\delta \right), & \text{otherwise},
\end{cases}
\end{equation}
where $x$ denotes the prediction error and $\delta$ is a threshold parameter that controls the transition between quadratic and linear regimes~\cite{huber1992robust}. In our implementation, we set the Huber threshold to $\delta = 0.1$, as summarized in Table~\ref{tab:algo_params}. By behaving quadratically for small errors and linearly for large errors, the Huber loss reduces sensitivity to outliers while preserving stable gradients for accurate estimates.

\begin{figure}[t]
  \centering
  \includegraphics[width=\linewidth]{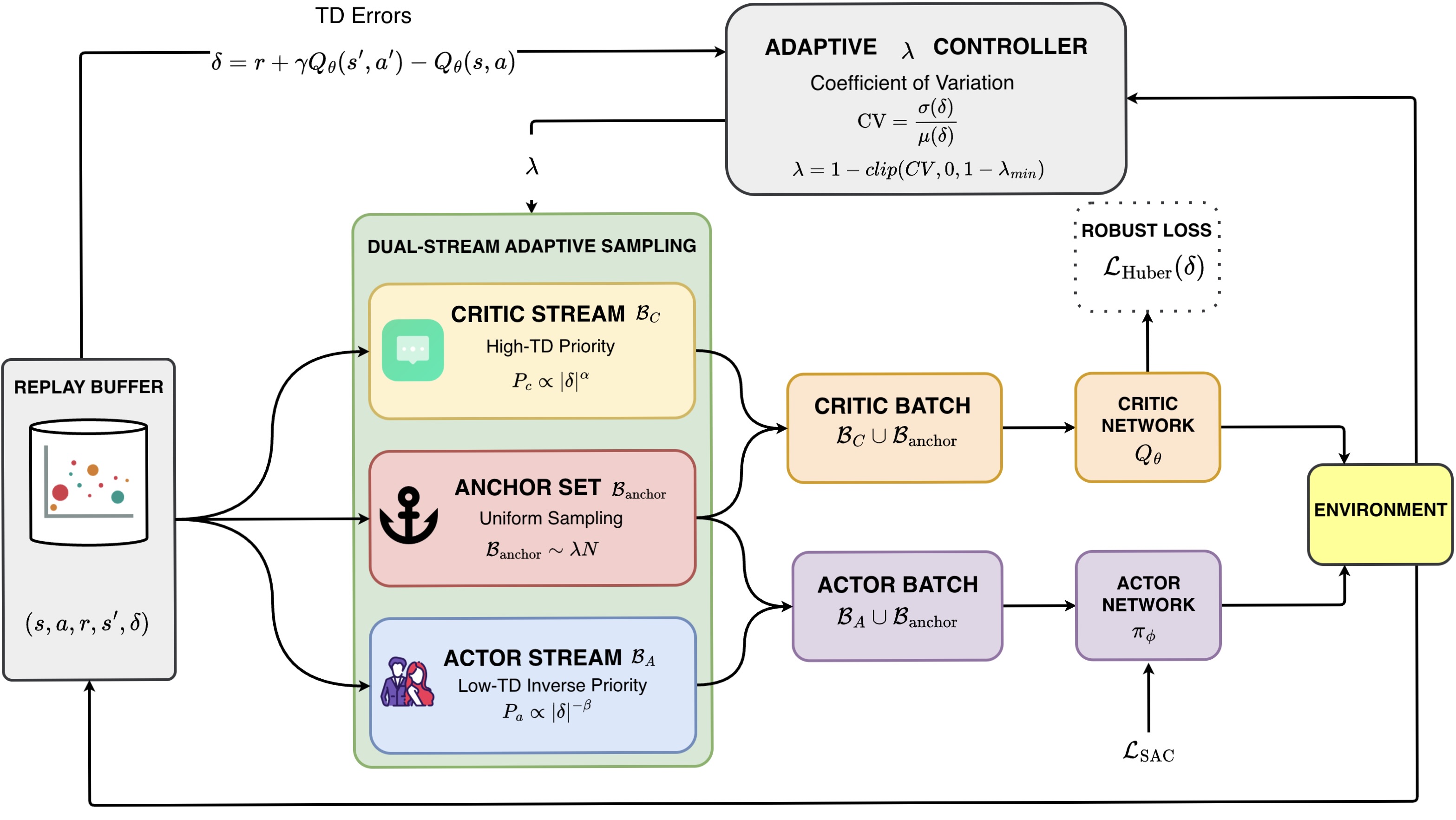}
  \caption{Overview of the proposed D-SPEAR: Dual-Stream Prioritized Experience
Adaptive Replay (D-SPEAR).
  The replay buffer is decomposed into an anchor set sampled uniformly and two prioritized streams:
  a high-TD stream for critic updates and a low-TD inverse-priority stream for actor updates.
  An adaptive controller adjusts the anchor ratio $\lambda$ based on the coefficient of variation (CV) of TD errors.}
  \label{fig:dspear_overview}
\end{figure}

Although experience replay, prioritization, and robust objectives have each been studied in isolation, their interaction in complex robotic manipulation settings remains insufficiently explored. As illustrated in Figure~\ref{fig:dspear_overview}, explicitly decoupling replay strategies for the actor and the critic provides a structured approach to addressing these challenges, which directly motivates the framework proposed in this work.

\section{Methodology: D-SPEAR: Dual-Stream Prioritized Experience Adaptive Replay}

\subsection{Actor--Critic Mismatch in Robotic Manipulation}

We consider robotic manipulation tasks modeled as Markov Decision Processes and optimized using
off-policy actor--critic algorithms. Let $Q_\theta(s,a)$ denote the critic parameterized by $\theta$,
and let $\pi_\phi(a|s)$ denote the stochastic policy parameterized by $\phi$.
In this setting, the critic provides value gradients that directly guide policy updates.

In contact-rich manipulation tasks, such as object lifting and articulated door opening,
large temporal-difference (TD) errors frequently arise near contact transitions, failure states,
or boundary conditions~\cite{APER2023actor}. Although such transitions are highly informative for
value learning, they often correspond to regions where the critic’s estimates are unreliable.
Updating the actor using gradients derived from these inaccurate value estimates introduces
high variance and bias into policy updates, which can lead to oscillatory behavior or
late-stage performance collapse.

This observation highlights a fundamental mismatch in the optimization objectives of the
actor--critic architecture: the critic benefits from replaying high-error transitions to
rapidly correct value estimates, whereas the actor requires stable and accurate value gradients
to ensure monotonic policy improvement. Applying identical replay strategies to both components
therefore induces conflicting optimization pressures and destabilizes training. This analysis provides an intuitive explanation of the actor–critic mismatch rather than a formal theoretical bound.

\subsection{Dual-Stream Experience Replay}

To address this mismatch, D-SPEAR decouples the replay sampling strategies for the actor and
the critic while maintaining a shared replay buffer $\mathcal{D}$.
Each training batch is composed of three elements: a uniformly sampled anchor set,
a critic-specific prioritized stream, and an actor-specific low-error stream.

\paragraph{Anchor Sampling}
To preserve distributional coverage and prevent excessive sampling bias, we sample an anchor
subset $\mathcal{B}_{\text{anchor}}$ uniformly from the replay buffer:
\begin{equation}
\mathcal{B}_{\text{anchor}} \sim \text{Uniform}(\mathcal{D}),
\end{equation}
with size $\lfloor \lambda N \rfloor$, where $N$ denotes the batch size and
$\lambda \in [0,1]$ controls the proportion of uniformly sampled transitions.
In practice, we construct a candidate set whose size is four times the batch size
(candidate sampling ratio = 4) before applying the prioritized sampling strategies.

\paragraph{Critic Stream}
The critic-specific subset $\mathcal{B}_C$ is sampled from the remaining transitions using
prioritized replay based on TD-error magnitude:
\begin{equation}
P_C(i) \propto |\delta_i|^{\alpha_c},
\end{equation}
where $\delta_i$ denotes the TD error of transition $i$, and $\alpha_c$ controls the strength
of prioritization. This stream emphasizes informative transitions to accelerate value convergence.
In our implementation, we set $\alpha_c = 1$, which corresponds to the standard PER prioritization strength.

\paragraph{Actor Stream}
The actor-specific subset $\mathcal{B}_A$ is sampled using an inverse-priority distribution
that favors low-error transitions:
\begin{equation}
P_A(i) \propto (|\delta_i| + \varepsilon)^{-\beta_a},
\end{equation}
where $\beta_a$ controls the preference for stable transitions and $\varepsilon > 0$
is a small constant introduced to prevent numerical instability when $|\delta_i| = 0$.
In our implementation, we set $\beta_a = 1$, which provides a balanced trade-off between
prioritization and distributional diversity. By updating the policy primarily using low-error
samples, the actor avoids unreliable gradients and achieves more stable policy improvement.

\subsection{Adaptive Anchor Mechanism}

A fixed anchor ratio $\lambda$ is suboptimal because the informativeness of TD-error
priorities changes throughout training. When TD errors exhibit strong relative
dispersion, prioritized replay provides a more discriminative learning signal.
By contrast, when TD errors are more homogeneous, the benefit of prioritization
diminishes and a larger uniformly sampled anchor subset helps maintain stable
state--action coverage.

To capture this effect, we adaptively adjust the anchor ratio $\lambda$, which
controls the fraction of uniformly sampled transitions in each training batch,
based on the coefficient of variation (CV) of TD-error magnitudes in the replay
buffer. At each update step, the CV is estimated from a fixed set of
$N_{\mathrm{CV}} = 1000$ TD-error samples drawn uniformly at random from the
replay buffer, independent of the training mini-batch size:
\begin{equation}
\text{CV} = \frac{\sigma_\delta}{\mu_\delta + \varepsilon}, \qquad
\lambda = 1 - \operatorname{clip}(\text{CV}, 0, 1 - \lambda_{\min}),
\end{equation}
where $\mu_\delta$ and $\sigma_\delta$ denote the mean and standard deviation
of the sampled TD-error magnitudes, respectively, $\varepsilon$ is a small
constant for numerical stability, and $\lambda_{\min}$ is the minimum anchor
ratio.

Intuitively, a larger CV indicates greater relative dispersion in TD errors,
so $\lambda$ decreases toward $\lambda_{\min}$ and the update places more
emphasis on prioritized replay. Conversely, when the CV is smaller, $\lambda$
increases and a larger fraction of uniformly sampled anchor transitions is
retained. In this way, the method automatically balances coverage and
prioritization during training without requiring manual task-specific tuning.

\subsection{Robust Critic Objective}

Robotic manipulation tasks often involve heterogeneous reward scales and occasional
outlier transitions. To make critic updates less sensitive to large Bellman residuals,
D-SPEAR replaces the mean squared error (MSE) critic loss used in standard SAC with
the Huber loss defined in the previous subsection, while keeping the SAC target
unchanged. Let $\mathcal{B}_c$ denote the critic replay batch. Then each critic is
optimized by
\begin{equation}
\mathcal{L}_Q(\theta_i)
=
\mathbb{E}_{\mathcal{B}_c}
\left[
\mathcal{L}_{\delta}\!\left(Q_{\theta_i}(s,a)-y\right)
\right],
\quad i \in \{1,2\},
\end{equation}
where
\begin{equation}
\begin{aligned}
y = r + \gamma(1-d)\Big(
&\min_{j\in\{1,2\}} Q_{\bar{\theta}_j}(s',a') \\
&- \alpha \log \pi_{\phi}(a' \mid s')
\Big),
\end{aligned}
\end{equation}
and $a' \sim \pi_{\phi}(\cdot \mid s')$. In all experiments, we set the Huber
threshold to $\delta = 0.1$.

Compared with MSE, this modification preserves the standard SAC Bellman target but
is intended to reduce the sensitivity of critic updates to rare large TD errors.

The overall training procedure of D-SPEAR integrates the dual-stream replay mechanism,
adaptive anchor adjustment, and robust critic optimization described above.
For clarity, Algorithm~\ref{alg:dspear} summarizes the complete learning process within
an off-policy actor--critic framework.

\begin{algorithm}[t]
    \caption{Dual-Stream Prioritized Experience Adaptive Replay (D-SPEAR)}
    \label{alg:dspear}
    \begin{algorithmic}[1]
        \REQUIRE Batch size $N$, minimum anchor ratio $\lambda_{\min}$, priority \\
        \hspace{1.1em} exponents $\alpha_c$, $\beta_a$, Huber threshold $\delta$
        \ENSURE Trained policy $\pi_\phi$
        
        \STATE Initialize policy network $\pi_\phi$, critic network $Q_\theta$, replay \\
        \quad buffer $\mathcal{D}$
        \FOR{each environment step $t$}
            \STATE Execute action $a_t \sim \pi_\phi(\cdot|s_t)$, observe $(r_t, s_{t+1})$
            \STATE Store transition $(s_t, a_t, r_t, s_{t+1})$ in $\mathcal{D}$ with maximum \\
            \quad priority
            \IF{$\mathcal{D}$ is ready for update}
                \STATE Sample TD errors from $\mathcal{D}$ to estimate $\mu_\delta, \sigma_\delta$
                \STATE Compute $CV \leftarrow \sigma_\delta / (\mu_\delta + \varepsilon)$
                \STATE Update anchor ratio $\lambda \leftarrow 1 - \text{clip}(CV, 0, 1 - \lambda_{\min})$
                \STATE Sample anchor set $\mathcal{B}_{\text{anchor}}$ of size $\lfloor \lambda N \rfloor$ uniformly from $\mathcal{D}$
                \STATE Sample critic indices $\mathcal{I}_C \sim |\delta_i|^{\alpha_c}$ of size $N - |\mathcal{B}_{\text{anchor}}|$
                \STATE Sample actor indices $\mathcal{I}_A \sim (|\delta_i| + \varepsilon)^{-\beta_a}$ of size $N - $\\
                \quad $|\mathcal{B}_{\text{anchor}}|$
                \STATE Construct batches $\mathcal{B}_C \leftarrow \mathcal{B}_{\text{anchor}} \cup \mathcal{I}_C$
                \STATE Construct batches $\mathcal{B}_A \leftarrow \mathcal{B}_{\text{anchor}} \cup \mathcal{I}_A$
                \STATE Update $\theta$ by minimizing Huber loss on $\mathcal{B}_C$:
                \STATE \quad $\theta \leftarrow \theta - \eta_Q \nabla_\theta \mathcal{L}_\delta(\mathcal{B}_C)$
                \STATE Update priorities in $\mathcal{D}$ using new TD errors from $\mathcal{B}_C$
                \STATE Update $\phi$ by maximizing SAC objective on $\mathcal{B}_A$:
                \STATE \quad $\phi \leftarrow \phi + \eta_\pi \nabla_\phi \mathbb{E}_{s \sim \mathcal{B}_A} [Q_\theta(s, \pi_\phi(s)) - $\\
                \quad $\alpha \log \pi_\phi(s)]$
                \STATE Update entropy temperature $\alpha$ and target networks
            \ENDIF
        \ENDFOR
    \end{algorithmic}
\end{algorithm}

\section{Experiment}

\subsection{Experimental Setup}

\begin{figure}[t]
    \centering
    \includegraphics[width=0.95\linewidth]{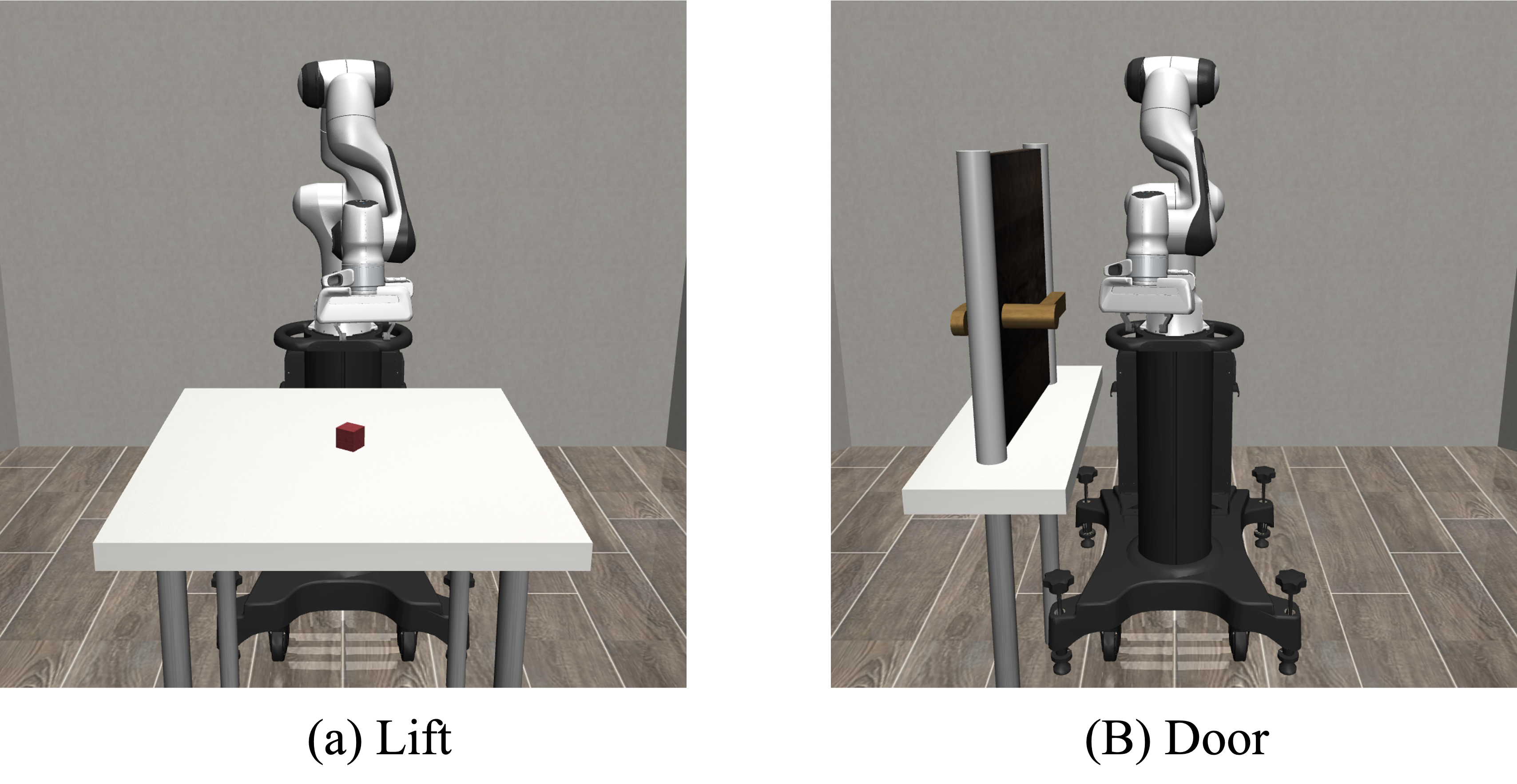}
    \caption{Robotic manipulation environments used in our experiments.
    (a) \textbf{Lift}: object grasping and lifting.
    (b) \textbf{Door}: contact-rich articulated door opening.}
    \label{fig:environment}
    \vspace{-1.5ex}
\end{figure}

\paragraph{Environments}
Experiments are conducted on two robotic manipulation tasks from the \textit{robosuite} framework: \texttt{Lift} and \texttt{Door}~\cite{robosuite2020}. 
Both tasks are instantiated with a 7-DoF Franka Panda robot and involve continuous state and action spaces with contact-rich dynamics. 
The \texttt{Lift} task requires the robot to grasp and lift an object from a table surface to a target height, whereas the \texttt{Door} task involves opening a hinged door through sustained interaction with its handle. 
These environments feature long-horizon decision making, delayed rewards, and sensitive contact transitions, making them well suited for evaluating training stability in robotic manipulation. Sample efficiency is evaluated using learning curves that report episode return as a function of environment interaction steps. Representative visualizations of both environments are shown in Figure~\ref{fig:environment}.

\paragraph{Environment Configuration}
All environments are controlled using an operational space controller in position mode (OSC\_POSITION). 
The control frequency is set to $20$ Hz, and each episode has a fixed horizon of $500$ time steps. 
Visual observations are disabled, and only low-dimensional proprioceptive and object-state observations are used. 
Reward shaping provided by \textit{robosuite} is enabled to facilitate learning while preserving task difficulty.
A summary of the environment configuration is provided in Table~\ref{tab:env_params}.

\begin{table}[t]
\centering
\caption{Robotic manipulation environment and training setup parameters.}
\label{tab:env_params}
\begin{tabular}{l c}
\hline
\textbf{Parameter} & \textbf{Value} \\
\hline
Environment & Lift, Door \\
Robot & Franka Panda \\
Controller & OSC\_POSITION \\
Control frequency & 20 Hz \\
Episode horizon & 500 steps \\
Observation type & Low-dimensional state \\
Camera observations & Disabled \\
Reward shaping & Enabled \\
Action space & Continuous \\
\hline
Total training steps & $2.5 \times 10^{5}$ \\
Number of random seeds & 5 \\
\hline
\end{tabular}
\vspace{-1.5ex}
\end{table}

\paragraph{Algorithm Configuration}
D-SPEAR is implemented on top of a Soft Actor--Critic (SAC) backbone. 
Both the actor and critic networks are parameterized as two-layer multilayer perceptrons with $256$ hidden units per layer. 
A shared replay buffer stores transitions of the form $(s_t, a_t, r_t, s_{t+1}, d_t)$. 
The proposed dual-stream replay mechanism constructs separate mini-batches for actor and critic updates, as described in Section~III.
All algorithmic and training hyperparameters are summarized in Table~\ref{tab:algo_params}.

\begin{table}[t]
\centering
\caption{Algorithm and training hyperparameters.}
\label{tab:algo_params}
\begin{tabular}{l c}
\hline
\textbf{Parameter} & \textbf{Value} \\
\hline
Base algorithm & SAC \\
Actor/Critic network & 2-layer MLP (256 units) \\
Optimizer & Adam \\
Replay buffer & Shared \\
Batch size $N$ & 256 \\
Discount factor $\gamma$ & 0.99 \\
Warm-up steps & 5{,}000 \\
Updates per step & 1 \\
\hline
Minimum anchor ratio $\lambda_{\min}$ & 0.5 \\
Candidate sampling ratio & 4 \\
Critic priority exponent $\alpha_c$ & 1.0 \\
Actor inverse-priority temperature & 1.0 \\
Critic loss & Huber loss($\delta = 0.1$) \\
\hline
\end{tabular}
\end{table}

\paragraph{Training Protocol}
All methods are trained for $500$ episodes with a maximum of $500$ steps per episode, resulting in a total interaction budget of $2.5 \times 10^{5}$ environment steps. 
The first $5{,}000$ steps are used for random exploration before policy updates begin. 
After the warm-up phase, one gradient update is performed per environment step. 
Each experiment is repeated over $5$ random seeds. 
Performance is evaluated using episode return, reported as the mean across seeds.

\subsection{Main Results on Robotic Manipulation Tasks}

\begin{figure}[t]
    \centering
    \includegraphics[width=\columnwidth]{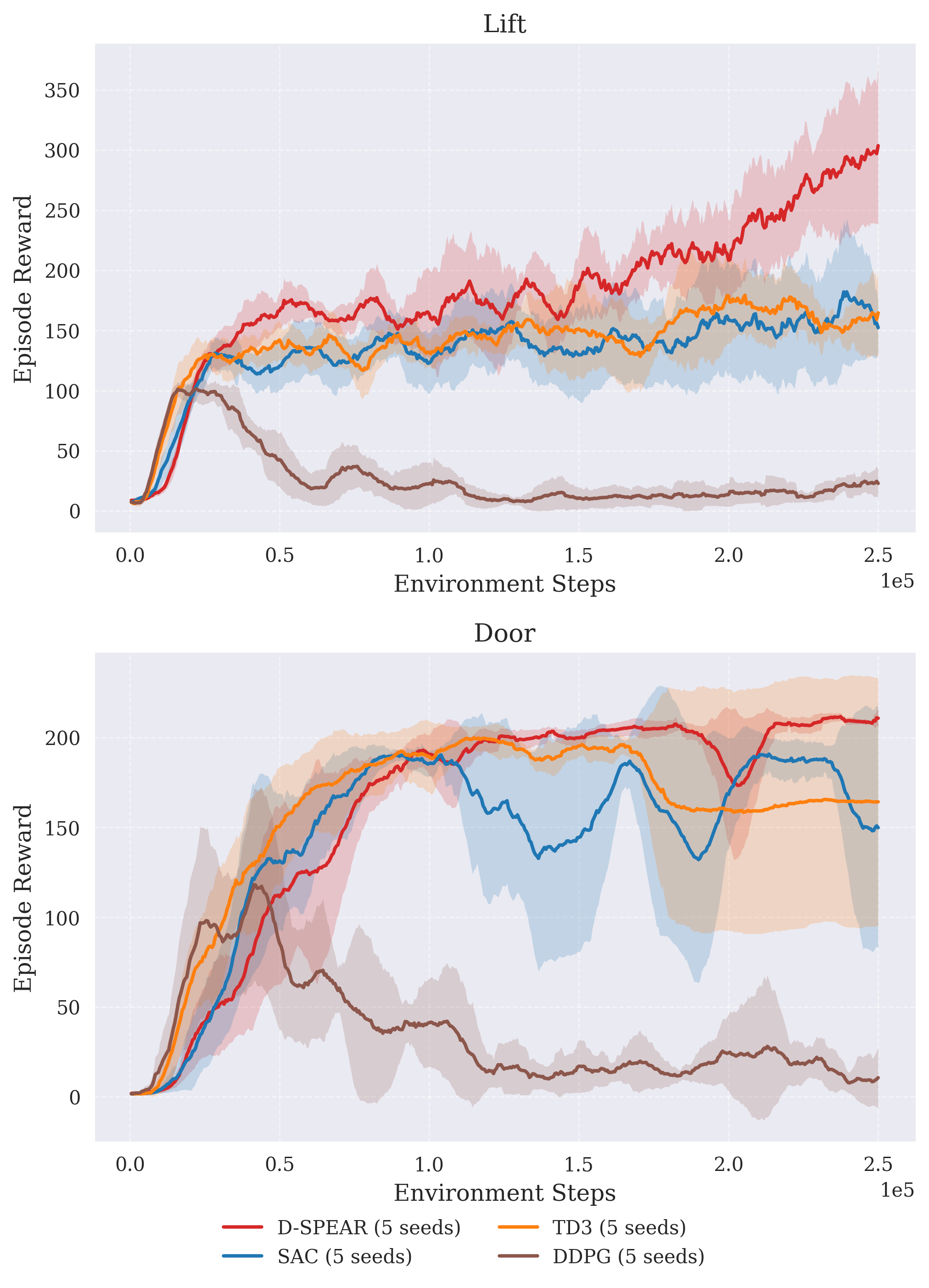}
    \caption{Performance comparison on robosuite manipulation tasks.
    Episode return as a function of environment steps on \textbf{Lift} and \textbf{Door}.
    Solid lines denote the mean performance over $5$ random seeds, and shaded regions indicate one standard deviation.}
    \label{fig:main_results}
    \vspace{-1.5ex}
\end{figure}

Figure~\ref{fig:main_results} compares D-SPEAR with standard off-policy baselines on the two robotic manipulation tasks. 
Across both environments, D-SPEAR consistently achieves superior final performance and exhibits substantially improved training stability.

\begin{table}[t]
\centering
\caption{Final performance on robosuite manipulation tasks.
Results are reported as the mean episode return over the final 10 evaluation episodes,
averaged across 5 random seeds.}
\label{tab:final_performance}
\begin{tabular}{lcccc}
\hline
\textbf{Environment} & \textbf{D-SPEAR} & \textbf{SAC} & \textbf{TD3} & \textbf{DDPG} \\
\hline
Lift  & \textbf{305.44} & 156.91 & 164.87 & 24.64 \\
Door  & \textbf{210.37} & 149.27 & 164.40 & 11.35 \\
\hline
\end{tabular}

\vspace{-1.5ex}
\end{table}

In addition to the learning curves in Figure~\ref{fig:main_results}, 
Table~\ref{tab:final_performance} summarizes the final performance of all methods. 
D-SPEAR achieves the highest final returns on both manipulation tasks, outperforming all baselines by a substantial margin. 
The performance gap is particularly pronounced on the \texttt{Lift} task, where stable long-horizon value estimation is critical.

\paragraph{Lift Task}
As shown in Figure~\ref{fig:main_results} and Table~\ref{tab:final_performance}, 
D-SPEAR demonstrates sustained performance improvement throughout training and attains the highest final episode return. 
In contrast, SAC and TD3 exhibit early learning progress but plateau at substantially lower performance levels, with noticeable variance across seeds. 
DDPG fails to maintain stable learning and collapses to near-zero returns after initial exploration.

\paragraph{Door Task}
The \texttt{Door} task presents a more challenging contact-rich manipulation problem due to articulated dynamics and delayed task completion signals. 
Figure~\ref{fig:main_results} shows that while SAC and TD3 initially achieve competitive returns, both methods suffer from pronounced performance oscillations and late-stage degradation. 
In contrast, D-SPEAR maintains stable learning dynamics and consistently converges to higher returns with reduced variance.

\subsection{Ablation Study}

\begin{figure}[t]
    \centering
    \includegraphics[width=\columnwidth]{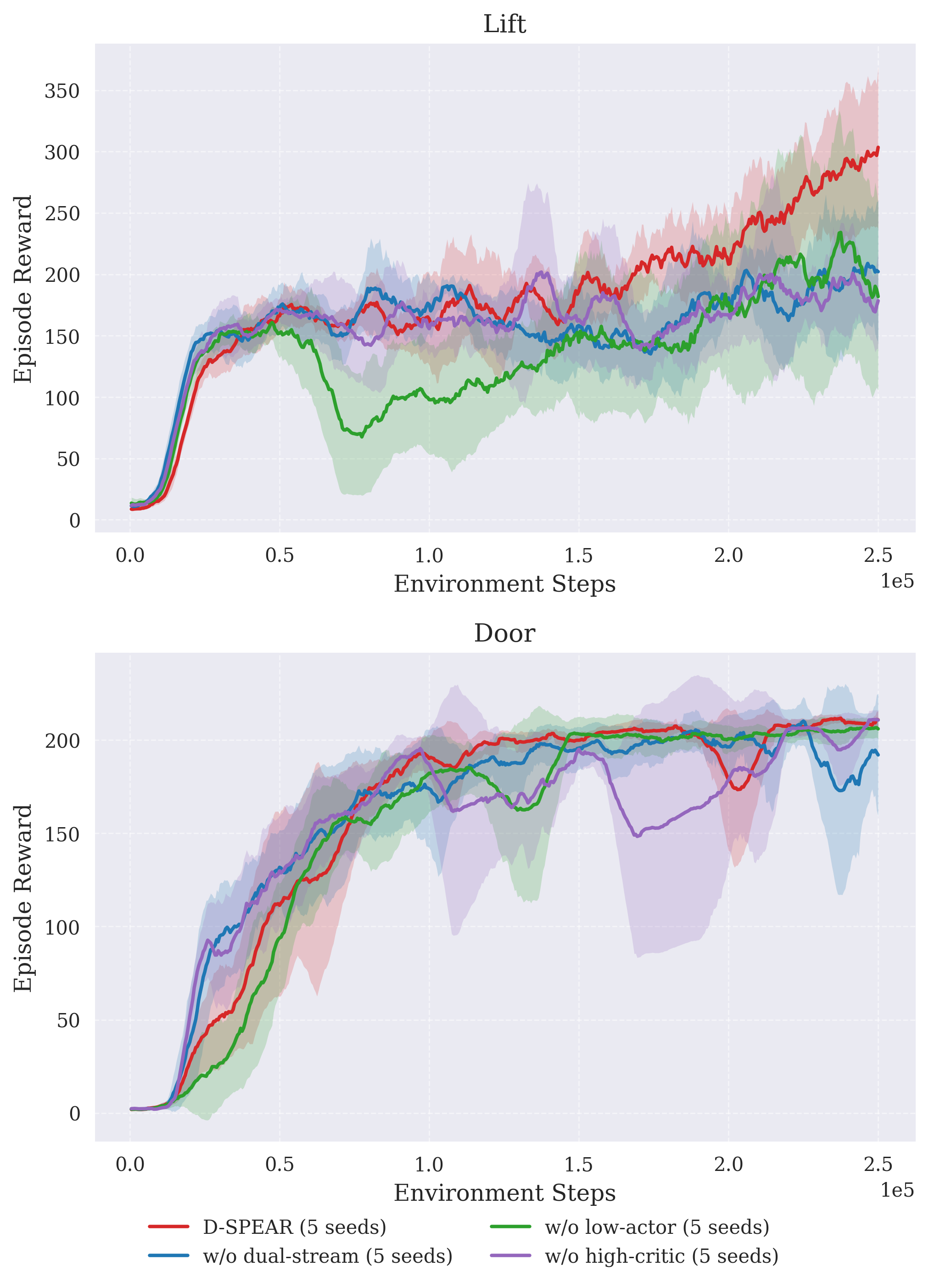}
    \caption{Ablation study on robosuite manipulation tasks.
    We compare the full D-SPEAR framework with variants that remove key components:
    \textbf{w/o dual-stream}, \textbf{w/o low-actor}, and \textbf{w/o high-critic}.
    Results are averaged over $5$ random seeds, with shaded regions indicating one standard deviation.}
    \label{fig:ablation_results}
    \vspace{-1.5ex}
\end{figure}

Figure~\ref{fig:ablation_results} presents the ablation results on both the \texttt{Lift} and \texttt{Door} tasks, 
illustrating the contribution of each component in D-SPEAR.

\paragraph{Effect of Dual-Stream Replay.}
As shown in Figure~\ref{fig:ablation_results}, removing the dual-stream structure (\textbf{w/o dual-stream}) 
significantly degrades performance on both tasks and increases training variance, closely resembling the behavior of the vanilla SAC baseline.

\paragraph{Effect of Critic-Side High-TD Sampling.}
Disabling critic-side high-TD prioritization (\textbf{w/o high-critic}) leads to slower convergence and pronounced oscillations, 
particularly in the \texttt{Door} task, as evidenced in Figure~\ref{fig:ablation_results}.

\paragraph{Effect of Actor-Side Low-TD Sampling.}
Removing the actor-side low-TD replay stream (\textbf{w/o low-actor}) results in unstable policy updates and intermittent performance degradation, 
highlighting the importance of constraining policy learning to reliable value estimates.

Overall, the ablation results in Figure~\ref{fig:ablation_results} demonstrate that the actor-side and critic-side replay streams in D-SPEAR play complementary roles. 
Their combined effect enables efficient value learning while suppressing policy instability, leading to robust performance across diverse robotic manipulation tasks.

\section{Conclusion and Future Work}

In this paper, we presented D-SPEAR, a dual-stream adaptive experience replay framework for off-policy actor--critic learning in robotic manipulation. By assigning separate replay streams to the critic and the actor, D-SPEAR addresses the mismatch between value estimation, which benefits from high-TD transitions, and policy improvement, which favors conservative updates. We combine this dual-stream design with a CV-based adaptive anchor mechanism and a robust critic objective to improve training stability.

We evaluated D-SPEAR on robosuite manipulation benchmarks using the Panda robot on the Lift and Door tasks. Across these settings, D-SPEAR achieved stronger final performance and lower variance than off-policy baselines, while ablations show that the actor-side and critic-side replay streams contribute to stable learning. These results highlight the benefit of decoupling replay strategies for actor and critic updates in contact-rich manipulation problems. By balancing prioritization and coverage through the adaptive anchor mechanism, the framework provides a practical approach for improving stability in off-policy reinforcement learning.

Future work will extend D-SPEAR in three directions. First, we will evaluate the method on a broader range of benchmark tasks, including additional robosuite environments and other manipulation suites to further assess generality. Second, we will investigate state-dependent and task-aware replay adaptation, for example by conditioning the anchor ratio or sampling distribution on task identity, episode phase, or state uncertainty instead of relying on a single global replay rule. Third, we will study real-robot and sim-to-real settings and explore integrating D-SPEAR with stronger actor--critic backbones and model-based components for scalable and reliable robot learning.

\bibliographystyle{IEEEtran}
\bibliography{reference}

@article{tang2025deep,
  title={Deep reinforcement learning for robotics: A survey of real-world successes},
  author={Tang, Chen and Abbatematteo, Ben and Hu, Jiaheng and Chandra, Rohan and Mart{\'\i}n-Mart{\'\i}n, Roberto and Stone, Peter},
  journal={Annual Review of Control, Robotics, and Autonomous Systems},
  volume={8},
  number={1},
  pages={153--188},
  year={2025},
  publisher={Annual Reviews}
}

@article{SAC1,
  title={Soft actor-critic algorithms and applications},
  author={Haarnoja, Tuomas and Zhou, Aurick and Hartikainen, Kristian and Tucker, George and Ha, Sehoon and Tan, Jie and Kumar, Vikash and Zhu, Henry and Gupta, Abhishek and Abbeel, Pieter and others},
  journal={arXiv preprint arXiv:1812.05905},
  year={2018}
}

@inproceedings{TD3,
  title={Addressing function approximation error in actor-critic methods},
  author={Fujimoto, Scott and Hoof, Herke and Meger, David},
  booktitle={International conference on machine learning},
  pages={1587--1596},
  year={2018},
  organization={PMLR}
}

@inproceedings{christmann2024benchmarking,
  title={Benchmarking Smoothness and Reducing High-Frequency Oscillations in Continuous Control Policies},
  author={Christmann, Guilherme and Luo, Ying-Sheng and Mandala, Hanjaya and Chen, Wei-Chao},
  booktitle={2024 IEEE/RSJ International Conference on Intelligent Robots and Systems (IROS)},
  pages={627--634},
  year={2024},
  organization={IEEE}
}

@article{APER2023actor,
  title={Actor prioritized experience replay},
  author={Saglam, Baturay and Mutlu, Furkan B and Cicek, Dogan C and Kozat, Suleyman S},
  journal={Journal of Artificial Intelligence Research},
  volume={78},
  pages={639--672},
  year={2023}
}

@inproceedings{robosuite2020,
  title={robosuite: A Modular Simulation Framework and Benchmark for Robot Learning},
  author={Yuke Zhu and Josiah Wong and Ajay Mandlekar and Roberto Mart\'{i}n-Mart\'{i}n and Abhishek Joshi and Kevin Lin and Soroush Nasiriany and Yifeng Zhu},
  booktitle={arXiv preprint arXiv:2009.12293},
  year={2020}
}

@article{Huber2020equivalence,
  title={An equivalence between loss functions and non-uniform sampling in experience replay},
  author={Fujimoto, Scott and Meger, David and Precup, Doina},
  journal={Advances in neural information processing systems},
  volume={33},
  pages={14219--14230},
  year={2020}
}

@inproceedings{SAC2,
  title={Soft actor-critic: Off-policy maximum entropy deep reinforcement learning with a stochastic actor},
  author={Haarnoja, Tuomas and Zhou, Aurick and Abbeel, Pieter and Levine, Sergey},
  booktitle={International conference on machine learning},
  pages={1861--1870},
  year={2018},
  organization={Pmlr}
}

@article{ER,
  title={Self-improving reactive agents based on reinforcement learning, planning and teaching},
  author={Lin, Long-Ji},
  journal={Machine learning},
  volume={8},
  number={3},
  pages={293--321},
  year={1992},
  publisher={Springer}
}

@inproceedings{PER,
  title     = {Prioritized Experience Replay},
  author    = {Schaul, Tom and Quan, John and Antonoglou, Ioannis and Silver, David},
  booktitle = {International Conference on Learning Representations (ICLR)},
  year      = {2016},
  note      = {Published as a conference paper},
  url       = {https://arxiv.org/abs/1511.05952}
}

@inproceedings{DDPG,
  title     = {Continuous Control with Deep Reinforcement Learning},
  author    = {Lillicrap, Timothy P. and Hunt, Jonathan J. and Pritzel, Alexander and Heess, Nicolas and Erez, Tom and Tassa, Yuval and Silver, David and Wierstra, Daan},
  booktitle = {International Conference on Learning Representations (ICLR)},
  year      = {2016},
  note      = {Published as a conference paper}
}

@ARTICLE{ACarchi,
  author={Grondman, Ivo and Busoniu, Lucian and Lopes, Gabriel A. D. and Babuska, Robert},
  journal={IEEE Transactions on Systems, Man, and Cybernetics, Part C (Applications and Reviews)}, 
  title={A Survey of Actor-Critic Reinforcement Learning: Standard and Natural Policy Gradients}, 
  year={2012},
  volume={42},
  number={6},
  pages={1291-1307},
  doi={10.1109/TSMCC.2012.2218595}}

@ARTICLE{10474399,
  author={Gao, Jian and Li, Yufeng and Chen, Yimin and He, Yaozhen and Guo, Jingwei},
  journal={IEEE Transactions on Instrumentation and Measurement}, 
  title={An Improved SAC-Based Deep Reinforcement Learning Framework for Collaborative Pushing and Grasping in Underwater Environments}, 
  year={2024},
  volume={73},
  number={},
  pages={1-14},
  doi={10.1109/TIM.2024.3379048}}

@article{levine2016end,
  title={End-to-end training of deep visuomotor policies},
  author={Levine, Sergey and Finn, Chelsea and Darrell, Trevor and Abbeel, Pieter},
  journal={Journal of Machine Learning Research},
  volume={17},
  number={39},
  pages={1--40},
  year={2016}
}

@inproceedings{kalashnikov2018scalable,
  title={Scalable deep reinforcement learning for vision-based robotic manipulation},
  author={Kalashnikov, Dmitry and Irpan, Alex and Pastor, Peter and Ibarz, Julian and Herzog, Alexander and Jang, Eric and Quillen, Deirdre and Holly, Ethan and Kalakrishnan, Mrinal and Vanhoucke, Vincent and others},
  booktitle={Conference on robot learning},
  pages={651--673},
  year={2018},
  organization={PMLR}
}

@incollection{huber1992robust,
  title={Robust estimation of a location parameter},
  author={Huber, Peter J},
  booktitle={Breakthroughs in statistics: Methodology and distribution},
  pages={492--518},
  year={1992},
  publisher={Springer}
}

\end{document}